\documentclass[letterpaper, 10 pt, conference]{ieeeconf}  
\usepackage{graphicx}
\usepackage{tabularx,booktabs}
\usepackage{array,multirow,graphicx}
\usepackage{makecell}
\usepackage{pdfpages}
\usepackage{amsmath}
\usepackage{mathabx}
\usepackage{booktabs} 
\usepackage{float}
\usepackage{gensymb}
\usepackage{float}

\IEEEoverridecommandlockouts                              

\title{\LARGE \bf
BAMF-SLAM: Bundle Adjusted Multi-Fisheye Visual-Inertial SLAM Using Recurrent Field Transforms
}

\author{Wei Zhang$^{1,2}$, Sen Wang$^{2,3}$, Xingliang Dong$^{4}$, Rongwei Guo$^{4}$ and Norbert Haala$^{1}$
\thanks{$^{1}$Institute for Photogrammetry, University of Stuttgart, Germany
        {\tt\small wei.zhang, norbert.haala@ifp.uni-stuttgart.de}}%
\thanks{$^{2}$Audiovisual Lab, Huawei Munich Research Center, Germany
        {\tt\small wei.zhang3, sen.wang@huawei.com}}%
\thanks{$^{3}$CAMP, Technical University of Munich, Germany
        {\tt\small sen.wang@tum.com}}%
\thanks{$^{4}$Central Media Technology Institute, Huawei 2012 Laboratories, China
        {\tt\small dongxingliang, guorongwei@huawei.com}}%
}

\begin{document}

\maketitle
\thispagestyle{empty}
\pagestyle{empty}

\begin{abstract}

In this paper, we present BAMF-SLAM, a novel multi-fisheye visual-inertial SLAM system that utilizes Bundle Adjustment (BA) and recurrent field transforms (RFT) to achieve accurate and robust state estimation in challenging scenarios. First, our system directly operates on raw fisheye images, enabling us to fully exploit the wide Field-of-View (FoV) of fisheye cameras. Second, to overcome the low-texture challenge, we explore the tightly-coupled integration of multi-camera inputs and complementary inertial measurements via a unified factor graph and jointly optimize the poses and dense depth maps. Third, for global consistency, the wide FoV of the fisheye camera allows the system to find more potential loop closures, and powered by the broad convergence basin of RFT, our system can perform very wide baseline loop closing with little overlap. Furthermore, we introduce a semi-pose-graph BA method to avoid the expensive full global BA. By combining relative pose factors with loop closure factors, the global states can be adjusted efficiently with modest memory footprint while maintaining high accuracy. Evaluations on TUM-VI, Hilti-Oxford and Newer College datasets show the superior performance of the proposed system over prior works. In the Hilti SLAM Challenge 2022, our VIO version achieves second place. In a subsequent submission, our complete system, including the global BA backend, outperforms the winning approach.

\end{abstract}

\vspace*{-1mm}
\section{INTRODUCTION}
\vspace*{-1mm}
Simultaneous Localization and Mapping (SLAM) is being widely deployed to real-world applications in various domains such as mobile robotics and augmented reality. In the active field of SLAM research, Visual-Inertial SLAM (VI-SLAM) has emerged as a highly robust and accurate solution. Despite significant progress in VI-SLAM research \cite{leutenegger2015keyframe,qin2018vins,campos2021orb,leutenegger2022okvis2}, emerging challenging datasets and new industrial applications \cite{schubert2018tum,zhang2022hilti} present new formidable challenges that demand increased accuracy and robustness from state-of-the-art VI-SLAM systems. For example, monitoring construction sites requires centimeter-level accuracy in low-texture environments with rapid lighting changes and motion. To address these challenges, we opt for the use of multiple fisheye cameras in combination with the complementary Inertial Measurement Unit (IMU) as sensor setup. This provides the system with sufficient sensing information, about both the environment and ego motion to handle challenging cases, resulting in more robust and efficient environment mapping.

\begin{figure}[t!]
\centering
\includegraphics[width=0.48\textwidth]{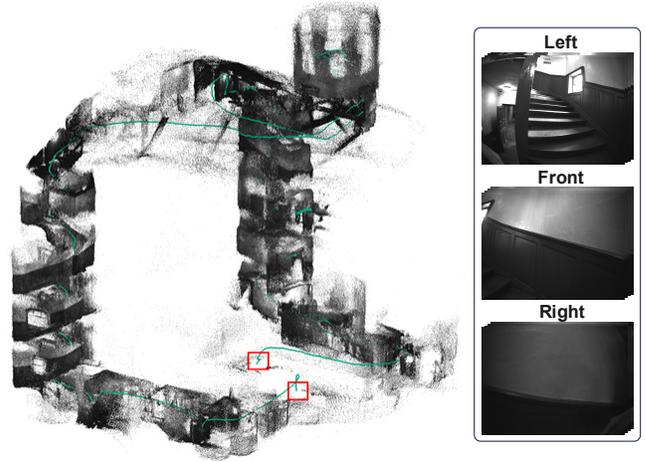}
\vspace*{-6mm}
\caption{Estimated point cloud and trajectory by the proposed system on exp09 sequence of Hilti-Oxford dataset. The start and end points of the trajectory are marked with red rectangles.}
\label{exp09}
\vspace*{-6mm}
\end{figure}

Bundle adjustment \cite{triggs2000bundle} has become the gold standard in modern SLAM systems. In feature-based approaches \cite{campos2021orb}, \cite{leutenegger2022okvis2}, the objective is to minimize the reprojection errors of the detected sparse features. In contrast, the photometric errors of dense pixels are minimized instead in direct approaches \cite{engel2014lsd}, \cite{engel2017direct}. With the increase of computational power and advances in deep learning, recent works \cite{teed2021droid,zhang2022towards,ji2022georefine} tend to combine the strengths of both worlds, which is to minimize the reprojection errors of dense depth maps using the prediction of RFT as targets. Based on this scheme, we built our system and extended it to a multi-camera VI-SLAM setup, which includes a unified joint factor graph formulation to fuse the inputs of multi cameras and IMU measurements in a tightly-coupled manner.

One major limitation of most visual SLAM systems is its reliance on visual appearance inside the camera view. They struggle to track the camera pose when the camera view is obstructed, contains too little texture, or has low overlap due to rapid movements. The great advantage of a multi-fisheye VI-SLAM system is that, on the one hand, the large FoV of fisheye cameras reduces the likelihood of low textures and increases the view overlap. When one view is fully obstructed, the other cameras or the IMU measurements can still provide redundancy and prevent the system from tracking failure or catastrophic drifting.

For global consistent mapping, we propose an approach called semi-pose-graph BA to jointly optimize poses and the depth maps involved by detected loop closures. Compared to the full BA, our method does not rely on reprojection factors to connect all keyframes, but instead converts the reprojection factors computed by the frontend into relative pose edges to build a pose graph. In contrast to pose graph optimization (PGO), our method optimizes not only the poses but also the depth maps of keyframes involved in detected loop closures, which are represented by reprojection factors. Thus our method is more accurate than PGO while also being more efficient than full BA.

In summary the main contributions of this work are:
\begin{itemize}
	\item{A state-of-the-art multi-fisheye VI-SLAM system called BAMF-SLAM that utilizes full image information from multiple fisheye cameras as well as complementary IMU measurements.}
    \item{A unified factor graph formulation that tightly-couples multi-camera inputs and IMU measurements to jointly optimize poses and depth maps.}
	\item{A semi-pose-graph bundle adjustment scheme that leverages relative pose edges and loop closures as reprojection factors to achieve efficient loop closing while maximizing global consistency.}
\end{itemize}

\section{RELATED WORK}
Both visual-inertial SLAM and multi-camera SLAM have received extensive research. To fuse visual and inertial measurements, most works, including this one, use the preintegration method of \cite{forster2015imu} to reduce the number of variables and efficiently solve the bundle adjustment problem. For a fixed computation cost of frontend, the marginalization of old states \cite{leutenegger2015keyframe} has been popular to model their uncertainty, whereas another line of works \cite{campos2021orb,mur2017visual} includes the old states as fixed prior in local BA. This work adopts the latter approach due to its simplicity and effectiveness.

Several works have explored the use of multiple cameras in SLAM systems. \cite{kuo2020redesigning} proposes a general design for multi-camera SLAM system including an initialization strategy and keyframe selection. MultiCol \cite{urban2017multicol} extends the ORB-SLAM2 \cite{mur2017orb} for the use with multiple fisheye camera inputs via bundle adjustment. \cite{ji2020panoramic} proposes a panoramic camera model for combining the images of a multi-camera rig, but only applicable when the camera centers have a slight offset. Our sensor configuration does not meet the requirement and thus cannot use the panoramic model. 

While there have been a few previous attempts to combine multi-camera inputs and inertial measurements, most methods rely on feature-based tracking across cameras. For instance, VILENS-MC \cite{zhang2021balancing} leverages cross-camera feature tracking to exploit camera overlaps. In contrast, our system does not track feature points, but instead establishes dense pixel-wise correspondences based on depth map representation. Furthermore, our system includes a global optimization backend resulting in further improvements in accuracy and robustness compared to previous works.

\section{Preliminary on Recurrent Field Transforms}
\begin{figure}[!htbp]
\centering
\includegraphics[width=0.43\textwidth]{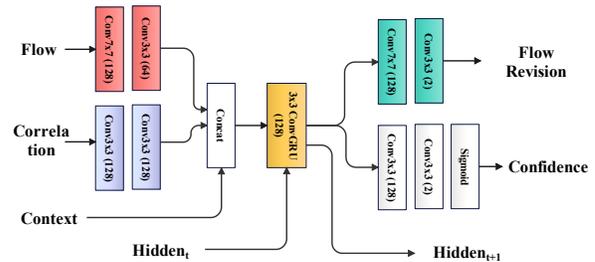}
\vspace*{-3mm}
\caption{Network structure for the update operation of recurrent field transforms borrowed from \cite{teed2021droid}.}
\label{gru}
\vspace*{-6mm}
\end{figure}

The concept of recurrent field transforms can be traced back to RAFT \cite{teed2020raft}, a deep network architecture for optical flow estimation. Inspired by this success, RAFT-3D \cite{teed2021raft3d} and RAFT-Stereo \cite{lipson2021raft} adopted this concept for the tasks of scene flow estimation and stereo matching respectively. Later, the concept of RFT was adopted for the dense SLAM system, DROID-SLAM \cite{teed2021droid}, which combines RFT with BA to solve poses and depths from optical flows. RFT predicts the dense pixel-wise correspondences between co-visible views based on the feature maps extracted from the input images using a residual-style network. The feature correlation can be looked up using the optical flow derived from the latest pose and depth estimates. As depicted in Fig. \ref{gru}, the inputs to the GRU module include, the current correlation, the context features, the last flow update, and the last hidden state. They are concatenated together to compute the update of the next flow revision and the associated confidence values.

To improve RFT-BA cooperation, DROID-SLAM enhances the training process by incorporating BA as a differential layer, allowing the loss function to be directed at pose and depth map estimates rather than optical flow and achieves the state-of-the-art accuracy on multiple benchmarks. This work builds based on DROID-SLAM and extends it to a multi-camera VI-SLAM system to deal with more challenging scenarios encountered in real-world applications.

Despite only being trained on the dataset \cite{wang2020tartanair} of pinhole images, the pre-trained model by \cite{teed2021droid} has excellent generalization ability on fisheye images. Therefore, we use the pre-trained model without further finetuning.

\vspace*{-1mm}
\section{System Overview}
\begin{figure}[thpb]
\centering
\vspace*{-3mm}
\includegraphics[width=0.48\textwidth]{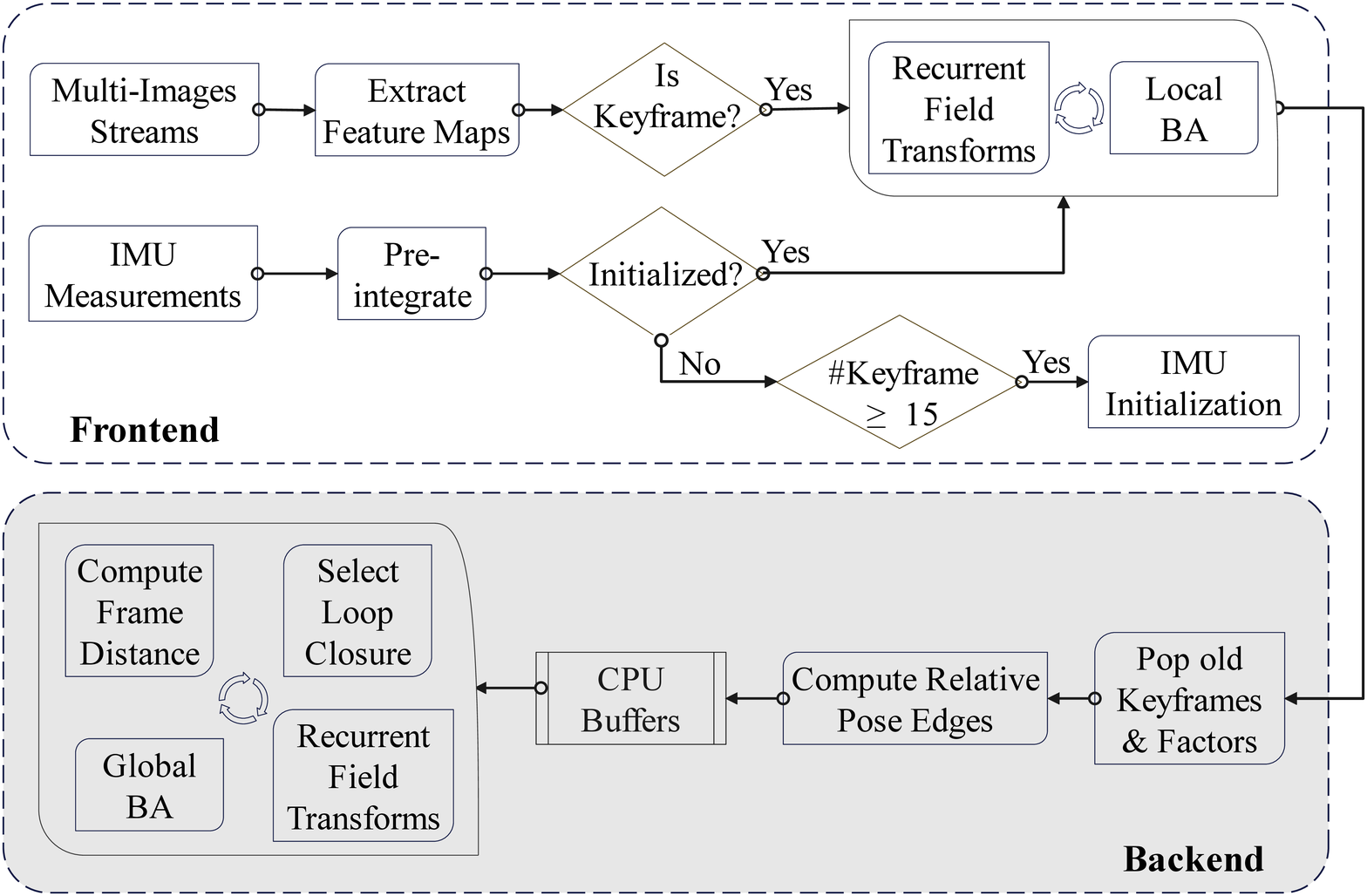}
\vspace*{-6mm}
\caption{Overview of the proposed system.}
\label{overview}
\vspace*{-4mm}
\end{figure}
The task of our proposed system is to estimate the state of the sensor platform as accurately and robustly as possible by utilizing the onboard multi-fisheye cameras and an IMU sensor. Figure \ref{overview} depicts a high-level overview of our system. Following the design of modern SLAM frameworks, we divide our system into frontend and backend. The frontend is responsible for processing the input images and IMU measurements, determining which are keyframes, keeping track of the most recent keyframes in a sliding window, and performing alternative RFT and local BA optimization. On the backend, it manages a data buffer of all keyframes established by the frontend and converts the intermediate results of reprojection factors into relative pose factors to build a global pose graph. Finally, to improve overall accuracy and global consistency, the backend employs loop closure detection and global semi-pose-graph BA.

\vspace*{-1mm}
\section{Method}
This section describes the proposed system's methodology in detail. We begin by discussing the application of the fisheye camera model to leverage the wide fisheye FoV in Sec. \ref{sec:fisheye}. Sec. \ref{sec:imu} covers the crucial IMU initialization process, which enables our system to combine visual-inertial information. Next, we present the factor graph formulation and explain each type of factors that are employed for fusing multiple inputs in Sec. \ref{sec:graph}. Finally, we introduce our proposed semi-pose-graph BA method (Sec. \ref{sec:global}) that achieves global consistency and improved accuracy by identifying long-term loop closures and maximizing the agreement between relative pose and loop closure constraints.

\vspace*{-1mm}
\subsection{Fisheye Camera Model}\label{sec:fisheye}
Conventional SLAM methods often assume an ideal pinhole camera model, which requires pre-processing of input images to remove distortion. However, when it comes to fisheye images, this can result in a loss of useful information due to cropped borders or sampling artifacts caused by the interpolation of objects at the edges. This negates the benefits of the wide FoV that fisheye images offer. To address this issue, our proposed system uses the Kannala-Brandt fisheye camera model \cite{kannala2006generic} as in \cite{campos2021orb} throughout our framework. The model is used for image pixel projection and unprojection, as well as the analytical computation of the Jacobians in the (uni-)projection functions. Unlike \cite{campos2021orb}, we use depth maps instead of sparse features and implement a custom CUDA kernel to parallelize the operation for all pixels.

\vspace*{-1mm}
\subsection{IMU Initialization}\label{sec:imu}
To enable the smooth integration of IMU observations, a set of new states that are not estimated in vision-only system need to be initialized, including gravity direction, scale, body velocity and biases of gyroscope and accelerometer. Following the best practice in \cite{campos2021orb}, it is preferable to start with vision-only BA for a short sequence with sufficient motion before initializing IMU. To summarize, our IMU initialization procedure consists of the following four steps:
\begin{enumerate}
	\item{Vision-only BA to establish the initial poses of the first $K$ keyframes.}
    \item{(Optional for static start) Least-square optimization for an initial gravity direction $R_{wg}$ and accelerometer bias $b_a$ using the constraint $\bar{g} - b_a = R_{wg} (0,0,G)^T$, where $G$ is the gravity magnitude and $\bar{g}$ is obtained by averaging the accelerometer measurements at standstill.}
	\item{Inertial-only BA to jointly optimize velocity, scale, biases $b_g$ as well as refine the estimates of the last two steps, namely $R_{wg}$ and $b_a$ using the preintegrated IMU factors (See Sec. \ref{sec:graph}).}
	\item{Visual-Inertial BA to finally optimize all variables using both visual reprojection and IMU preintegration factors.}
\end{enumerate}

In step 2, we determine whether the initial phase is stationary by examining the magnitude of the gyroscope measurement. If the initial phase is indeed stationary, the accelerometer data in this phase can provide strong prior information about the gravity direction.

The scale can typically be observed directly through a known stereo baseline via offline calibration and does not require optimization as a variable. However, our empirical study found that multi-camera calibration often lacks accuracy, and the inclusion of a scale variable can reduce the impact of calibration errors.

\vspace*{-1mm}
\subsection{Factor Graph Formulation}\label{sec:graph}
\begin{figure*}[thpb]
\centering
\includegraphics[width=0.9\textwidth]{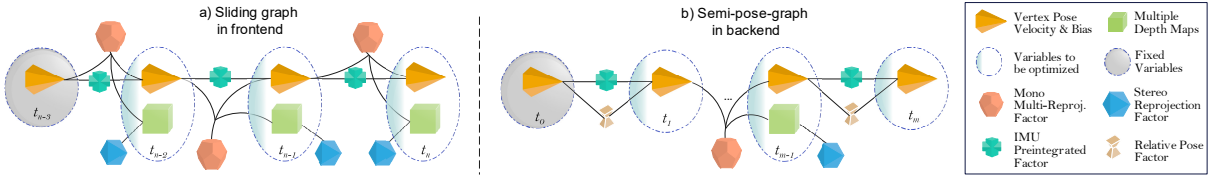}
\vspace*{-2mm}
\caption{Illustration of the proposed joint factor graph formulation. a)\,A sliding graph with a window size of three keyframes and one fixed keyframe served as prior. b)\,Semi-pose-graph with a long-term loop closure between two distant keyframes shown as an example.}
\label{graph}
\vspace*{-6mm}
\end{figure*}
The core to our bundle adjustment based system lies in the formulation of a factor graph that incorporates all types of factors to optimize the system states and 3D structure jointly. The 3D structure is represented by the depth maps of keyframes. The system state is defined in the IMU frame denoted by $B$ and the state at time $t_i$ can be defined as follows:
\begin{equation}\label{eq:state}
    \mathbf{x_{i}} = [\mathbf{T_{i}}, \mathbf{v_{i}}, \mathbf{b_{i}^{a}}, \mathbf{b_{i}^{g}}]^{T}
\end{equation}
where $\mathbf{T_{i}}=[\mathbf{R}_{i}, \mathbf{p}_{i}] \in$ SE(3) is the body pose, $\mathbf{v_{i}^{w}}$ is the velocity, and $\mathbf{b_{i}^{a}}$, $\mathbf{b_{i}^{g}}$ are the IMU accelerometer and gyroscope biases respectively. 

Fig. \ref{graph} depicts the factor graph formulations for the frontend and backend respectively. On the frontend, we maintain a sliding graph and optimize only the variables within a dynamic window. Upon a new keyframe, we first optimize the state of the incoming keyframe and its depth map, using a window size of 1, which we refer to as the tracking process. Next, the local BA is performed with a larger window size of $n$ keyframes. Through empirical observation, we have observed that directly running local BA on new keyframe requires more iterations to converge. However, if we perform local BA after tracking, which is significantly faster than local BA, it only takes two iterations to converge. This approach results in a significant improvement in runtime performance, as discussed in Sec. \ref{sec:runtime}. Moreover, a few vertices outside the window linked by reprojection factors are included in the optimization as fixed priors.

To incorporate IMU measurements, we utilize the IMU preintegration method \cite{forster2015imu} to compute IMU preintegration factors that connect consecutive keyframes. For reprojection factors, we distinguish between the mono and stereo types. Mono factors connect a pair of keyframes with common view, where vertex pose estimates are used to transform the coordinates during reprojection. In the stereo case, cross-camera reprojection is computed using the known camera extrinsics. Let $\Pi$ denote the fisheye projection function, a point $\mathbf{p}_i$ can be reprojected with an estimated depth $\hat{d}_i$ using the following equation:
\vspace*{-1mm}
\begin{equation}\label{equ:2}
	\mathbf{\hat{p}}_{ij} = \Pi(\mathbf{\hat{T}}_{ij} \Pi^{-1}(\mathbf{p}_i, \hat{d}_i))
\end{equation}
where $\mathbf{\hat{T}}_{ij} = T_{CB}\mathbf{\hat{T}}_{j}^{-1}\mathbf{\hat{T}}_{i}T_{CB}^{-1}$ refers to the relative transformation from keyframe $i$ to keyframe $j$ in the mono case. In the stereo case, it refers to the extrinsic $T_{C_i C_j}$ between camera $i$ and camera $j$.

For the reprojection targets, the RFT predicts the flow revision $\boldsymbol{\delta}_{ij}$ to add on the last optical flow $\mathbf{\hat{p}}_{ij}$. Thus, the updated optical flow, namely the dense correspondences, is obtained as follows:
\vspace*{-2mm}
\begin{equation}\label{equ:2}
	\widecheck{\mathbf{p}}_{ij} = \mathbf{\hat{p}}_{ij} + \boldsymbol{\delta}_{ij}
\end{equation}

To mitigate the impact of false correspondence and occlusion, the confidences $w_i$ is predicted for each pixel by RFT. Low confidences are assigned to pixels that do not have a counterpart in the target view or are ambiguous. The reprojection error then can be formulized as follows:
\vspace*{-1mm}
\begin{equation}\label{equ:3}
	r_{ij} = \| \widecheck{\mathbf{p}}_{ij} - \Pi(\mathbf{\hat{T}}_{ij} \Pi^{-1}(\mathbf{p}_i, \hat{d}_i)) \|_{w_i}^2
\end{equation}


As a special case in our multi-camera setup, one keyframe has multiple depth maps pointing to different view directions. In this case, the reprojection errors are computed for all depth maps, which we refer to as the multi-reprojection factor. For Newer College and Hilti-Oxford dataset, we include the depth maps of the left and right views in addition to the front view.


\vspace*{-1mm}
\subsection{Semi-Pose-Graph BA}\label{sec:global}
\begin{figure}[thpb]
\centering
\vspace*{-3mm}
\includegraphics[width=0.42\textwidth]{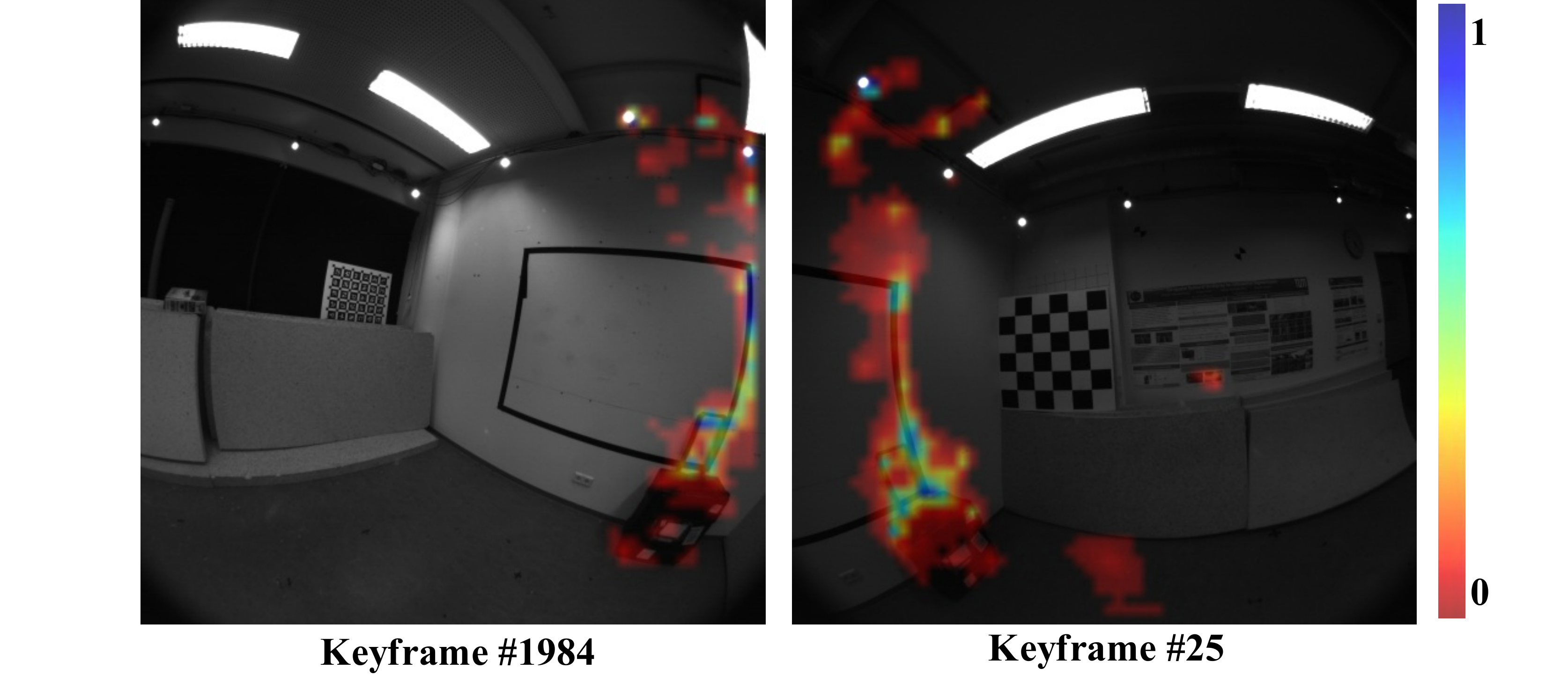}
\vspace*{-3mm}
\caption{Example of a wide baseline loop closure with color-coded confidence values predicted by the RFT network. Note that values lower than 0.01 are omitted for clarity.}
\label{lc}
\vspace*{-2mm}
\end{figure}

\begin{figure*}[t!]
\centering
\includegraphics[width=0.9\textwidth]{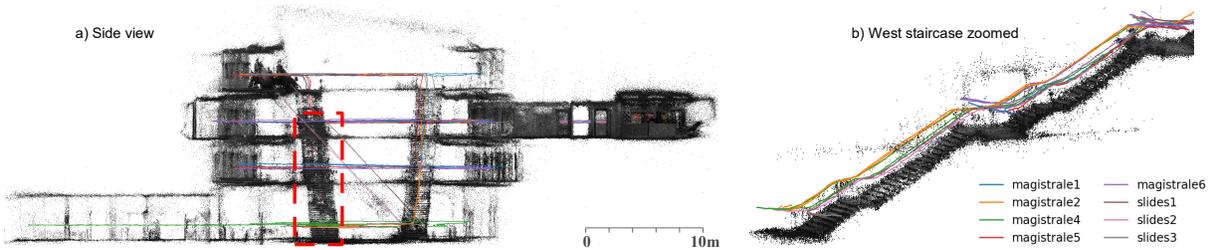}
\vspace*{-2mm}
\caption{Qualitative results on TUM-VI dataset. As groundtruth only covers a single room, we transform all results to the groundtruth frame for consistency check. Estimated trajectories are depicted in colors and point clouds are accumulated. Red-box marked region is zoomed in on the right from the side look to provide a better illustration. Notably our results demonstrate high consistency in cleanly separating floors and clear-cut edges of stairs.}
\label{swarm}
\vspace*{-6mm}
\end{figure*}
Performing global full BA is computationally expensive and requires a large memory footprint as it optimizes the system states and depth maps of all keyframes using all reprojection factors by the frontend. Additionally, since the frontend already provides states with high relative accuracy and local consistency, it may not be necessary to optimize the already very accurate relative pose. To this end, we have determined that the frontend's reprojection factors can be converted into relative pose factors using the dense correspondences established by local BA. A pose graph can then be constructed using these relative pose factors. The geometry constraints underlying the reprojection factors is primarily represented by the relative pose factors and their associated covariances. We compute these relative pose factors and covariances using the Gauss-Newton method \cite{niemeier2008ausgleichungsrechnung} as follows:
\vspace*{-1mm}
\begin{align}
    \vspace*{-2mm}
    \mathbf{H} \mathbf{\Delta x} &= \mathbf{J}^T \mathbf{W} \mathbf{J} \mathbf{\Delta x} = \mathbf{b} \\
    \mathbf{Cov} &= (\mathbf{J} \mathbf{\Delta x} - \mathbf{r})^T \mathbf{W} (\mathbf{J} \mathbf{\Delta x} - \mathbf{r}) \mathbf{H}^{-1}
\end{align}
where $\mathbf{J}$ is the Jacobian of the reprojection function w.r.t the relative pose, $\mathbf{r}$ is the last reprojection error vector and $\mathbf{W}$ is the weight matrix composed by the confidence predicted by RFT. The update to the relative pose is denoted as $\mathbf{\Delta x}$.

The trajectory estimate by the frontend is prone to odometry drift, which accumulates over long distances traveled. To mitigate drift error, the backend searches for potential loop closures to add as additional constraints. The backend starts by computing the frame distances between all keyframes, which is determined by the mean optical flow calculated based on the last pose and depth map estimates. A relaxed frame distance threshold is typically used to allow for more potential loop closures with wider baselines. While this may lead to the inclusion of wrong loop closures, their impact is insignificant because the RFT network is highly reliable and assigns low confidence to the wrong matches. Fig. \ref{lc} depicts a wide baseline loop closure, where the same room is revisited after a long trip. The wide FoV of the fisheye camera used in our system enables the detection of such loop closures even with little overlap at the image border.

With the use of relative pose factors, a pose graph can be constructed.  In addition, the detected loop closures can be presented in the form of reprojection factors, which extend the pose graph into a semi-pose-graph that includes depth maps involved in loop closures, as shown in Fig. \ref{graph},b). The semi-pose-graph BA is more efficient than full BA, as the number of variables to be optimized is much smaller. Moreover, the relative pose factors, which are derived from the reprojection factors established by the frontend, are highly accurate and robust due to the significant overlap of keyframes within the frontend sliding window. As a result, the semi-pose-graph BA is both efficient and accurate.

\vspace*{-2mm}
\section{Experiments}
\vspace*{-1mm}
Extensive evaluations has been carried out on our proposed system using diverse datasets. To validate the effectiveness of our system in stereo+IMU setup, we evaluate in various indoor scenes from TUM-VI dataset \cite{schubert2018tum}. We further evaluate the multi-camera setup on the challenging Newer College dataset \cite{zhang2021multi}, which contains indoor and large-scale outdoor scenes, and the Hilti-Oxford dataset \cite{zhang2022hilti}, which includes challenging sequences collected on construction sites and a historical theatre. For all evaluations, both VIO and VI-SLAM results are provided to differentiate the performance in terms of odometry drift by frontend and overall performance with global optimization. Finally, we present an analysis of the processing time and demonstrate the potential of real-time capability.

\subsection{Implementation Details}
\vspace*{-1mm}
The framework is implemented in Python/CUDA using PyTorch and Eigen libraries, and tested on a desktop PC with an Intel i9-12900K CPU and an Nvidia RTX 3090Ti GPU. Input images are downsampled to 440$\times$440 and 384$\times$512 resolution for TUM-VI and the other two datasets respectively. To reduce the memory footprint, a ring buffer data structure is implemented for storing image feature maps and intermediate results of keyframes. This way, only a fixed amount of GPU memory needs to be allocated regardless of data length, and the data of each new incoming keyframe is written to the beginning of the buffer. Our system can run on GPUs with 11GB of memory for both the frontend and the backend, which ensures high performance even for large-scale datasets.

\subsection{Performance on TUM-VI Dataset}
\vspace*{-1mm}
\setlength{\tabcolsep}{0pt}
\begin{table}[t]
\caption{Absolute Trajectory Error (ATE) in meters on the TUM-VI dataset \cite{schubert2018tum}. Best results are in bold.}
\vspace*{-4mm}
\label{tum-vi}
\begin{center}
\resizebox{0.43\textwidth}{!}{
\begin{tabular}{lc|ccc|ccc|c}
\toprule
&& \multicolumn{3}{c|}{Stereo-VIO} & \multicolumn{3}{c|}{Stereo-VI-SLAM} & \\
\multicolumn{2}{c|}{Seq} & \ BASALT \ & \ OKVIS2 \ & \, \, Ours \, \, & \ \makecell{ORB-\\SLAM3} \ & \ OKVIS2 \ & \, \, Ours \, \, & \ \makecell{Length\\{[m]}} \\
\midrule
\multirow{6}{*}{\rotatebox[origin=c]{90}{corridor}} & 1 & 0.34 & 0.54 & \textbf{0.31} & 0.03 & 0.02 & \textbf{0.01} & 305\\
& 2 & 0.42 & 0.44 & \textbf{0.15} & 0.02 & 0.06 & \textbf{0.01} & 322\\
& 3 & 0.35 & 0.55 & \textbf{0.21} & 0.02 & 0.03 & \textbf{0.01} & 300\\
& 4 & 0.21 & 0.11 & \textbf{0.04} & 0.21 & 0.10 & \textbf{0.01} & 114\\
& 5 & 0.37 & 0.51 & \textbf{0.17} & \textbf{0.01} & 0.09 & \textbf{0.01} & 270\\
& avg \ & 0.34 & 0.43 & \textbf{0.18} & 0.06 & 0.06 & \textbf{0.01} & 262\\
\midrule
\multirow{6}{*}{\rotatebox[origin=c]{90}{magistrale}} & 1 & \textbf{1.20} & 2.03 & 1.74 & 0.24 & 0.07 & \textbf{0.01} & 918\\
& 2 & 1.11 & 3.22 & \textbf{1.00} & 0.52 & 1.22 & \textbf{0.01} & 561\\
& 3 & \textbf{0.74} & 2.11 & 1.53 & 1.86 & \textbf{0.09} & 1.39 & 566\\
& 4 & 1.58 & 1.94 & \textbf{1.26} & 0.16 & 0.25 & \textbf{0.01} & 688\\
& 5 & 0.60 & 1.01 & \textbf{0.21} & 1.13 & 0.02 & \textbf{0.01} & 458\\
& 6 & 3.23 & 2.32 & \textbf{1.66} & 0.97 & 0.76 & \textbf{0.01} & 771\\
& avg & 1.41 & 2.10 & \textbf{1.23} & 0.81 & 0.40 & \textbf{0.24} & 660\\
\midrule
\multirow{6}{*}{\rotatebox[origin=c]{90}{room}} & 1 & 0.09 & 0.06 & \textbf{0.02} & \textbf{0.01} & \textbf{0.01} & \textbf{0.01} & 146\\
& 2 & \textbf{0.07} & \textbf{0.07} & \textbf{0.07} & \textbf{0.01} & \textbf{0.01} & \textbf{0.01} & 142\\
& 3 & 0.13 & 0.06 & \textbf{0.04} & \textbf{0.01} & \textbf{0.01} & \textbf{0.01} & 135\\
& 4 & 0.05 & \textbf{0.02} & \textbf{0.02} & \textbf{0.01} & \textbf{0.01} & \textbf{0.01} & 68\\
& 5 & 0.13 & \textbf{0.02} & \textbf{0.02} & \textbf{0.01} & \textbf{0.01} & \textbf{0.01} & 131\\
& 6 & 0.02 & \textbf{0.02} & \textbf{0.02} & \textbf{0.01} & \textbf{0.01} & \textbf{0.01} & 67\\
& avg & 0.08 & 0.04 & \textbf{0.03} & \textbf{0.01} & \textbf{0.01} & \textbf{0.01} & 115\\
\midrule
\multirow{4}{*}{\rotatebox[origin=c]{90}{slides}} & 1 & \textbf{0.32} & 0.96 & 0.43 & 0.41 & 0.37 & \textbf{0.01} & 289\\
& 2 & \textbf{0.32} & 0.74 & 0.51 & 0.49 & 0.16 & \textbf{0.01} & 299\\
& 3 & 0.89 & 2.51 & \textbf{0.41} & 0.47 & 0.13 & \textbf{0.01} & 383\\ 
& avg & 0.51 & 1.40 & \textbf{0.45} & 0.45 & 0.22 & \textbf{0.01} & 324\\
\bottomrule
\end{tabular}}
\vspace*{-8mm}
\end{center}
\end{table}
The TUM-VI \cite{schubert2018tum} dataset was collected using a handheld device equipped with two fisheye cameras in a stereo setup and an IMU. We selected four challenging scenes, including low-texture (slides), rapid rotation (corridor), and large-scale multi-story building (magistrale), to evaluate our approach. Tab. \ref{tum-vi} summarizes the quantitative evaluation results and the comparison to the previous works. We compared our VIO setup against BASALT \cite{usenko2019visual} and the VIO version of OKVIS2 \cite{leutenegger2022okvis2}, both of which are feature-based VIO methods without loop closure. Our method outperforms all other methods in terms of averaged ATE across all four scenes.

For the evaluation on VI-SLAM systems, the trajectory optimized globally with loop closures is evaluated. Our full SLAM system outperforms both OKVIS2 \cite{leutenegger2022okvis2} and ORB-SLAM3\cite{campos2021orb}, achieving the lowest ATE. Notably, our system's ability to detect and close loop closures with little overlap allows us to reach an ATE of 1cm for almost all sequences. Fig. \ref{lc} shows a wide baseline loop closure from the sequence \emph{corridor 1}. On the sequence \emph{magistrale 3}, due to the larger odometry drift and thus a greater frame distance, loop closures cannot be successfully found.

Notably, the groundtruth is only available at the beginning and end of the sequences, limiting the groundtruth coverage to motion within a single room. Therefore, the ATE reported in Tab. \ref{tum-vi} do not fully represent the accuracy of the estimated trajectory outside of the room. To qualitatively evaluate the overall accuracy, we transform the results to the groundtruth frame using SE3 alignment and examine the consistency of the accumulated point clouds. Fig. \ref{swarm} shows the accumulated point clouds along with their respective trajectories, depicted in different color.

\subsection{Performance on Newer College Dataset}
\vspace*{-1mm}
\begin{table}[h]
\vspace*{-2mm}
\caption{Evaluations on Newer College Dataset based on Relative Pose Error (RPE) and ATE (Unit: Meter).}
\vspace*{-3mm}
\label{NCD}
\begin{center}
\resizebox{0.45\textwidth}{!}{
\begin{tabular}{c|cccc|cccc|c}
\toprule
& \multicolumn{4}{c|}{10m RPE} & \multicolumn{4}{c|}{ATE} & \\
\cmidrule{2-9}
\multirow{2}{*}{Seq} \ & Open & ORB- & VILENS & Ours & \multicolumn{2}{c}{Our VIO} & \multicolumn{2}{c|}{VI-SLAM} & \ \multirow{2}{*}{\makecell{Length\\{[m]}}} \ \\
& \ VINS \ & \ SLAM3 \ & \ MC \ & \ MC-VIO \               & \ Stereo \ & \, MC \, \, & \ \ Stereo \ & \, MC \, \, \\
\midrule
MATH    & 0.65 & Fail & 0.26 & \textbf{0.12} & 0.47 & 0.40 & 0.21 & \textbf{0.20} & 329 \\
MINE    & 0.98 & Fail & 0.20 & \textbf{0.07} & 0.28 & 0.45 & \textbf{0.08} & 0.11 & 236 \\
QUAD    & 1.01 & 0.23 & 0.31 & \textbf{0.18} & 0.73 & 0.64 & \textbf{0.24} & 0.37 & 244 \\
STAIRS \ & 0.33 & 0.20 & 0.16 & \textbf{0.09} & 0.09 & 0.07 & 0.05 & \textbf{0.04} & 59 \\
PARK \ & 0.18 & 0.13 & n/a & \textbf{0.10} & 4.27 & 5.08 & \textbf{0.53} & 1.04 & 2396 \\
\bottomrule
\end{tabular}}
\vspace*{-5mm}
\end{center}
\end{table}
The Newer College dataset \cite{zhang2021multi} includes four fisheye cameras and an IMU sensor. The difficulty and variety of the scenes range from narrow staircase, to dark underground mine, and over 2\,km outdoor walk. The groundtruth is provided for the entire sequence by aligning lidar scans to detailed prior maps. Tab. \ref{NCD} presents the evaluation results on the Newer College dataset. Following \cite{zhang2021balancing}, we compare our VIO system to OpenVINS \cite{geneva2020openvins}, ORB-SLAM3 \cite{campos2021orb} and VILENS-MC \cite{zhang2021balancing} based on the metric Relative Pose Error (RPE) over 10m. Our results demonstrate that our method outperforms all other methods on all scenarios. Notably, VILENS-MC, which also uses a multi-camera setup, is the closest method to our approach, highlighting the benefits of leveraging multi-camera inputs.
\begin{figure}[thpb]
\vspace*{-3mm}
\centering
\includegraphics[width=0.49\textwidth]{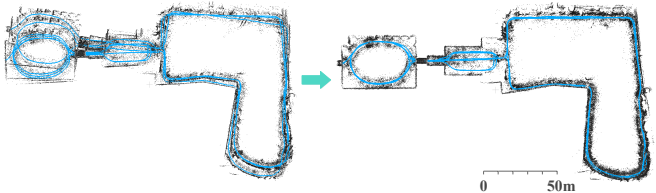}
\vspace*{-6mm}
\caption{Before and after global semi-pose-graph BA on sequence \emph{Park}.}
\label{park}
\vspace*{-2mm}
\end{figure}

In addition to using relative metrics, we also assess absolute accuracy using the ATE metric. Surprisingly, our experiments show that the error in the stereo setting is lower than in the multi-camera (MC) setting. We suspect this discrepancy may be due to errors in the extrinsic calibration and plan to investigate further in future work. When comparing the VIO and VI-SLAM results, the VI-SLAM results are significantly better than the VIO results thanks to the loop closing and semi-pose-graph BA. It could be argued that the VIO frontend has provided a fair good starting point for the backend, with relatively low odometry drift. Based on this, the backend was then able to identify potential loop closures by measuring frame distances between keyframes. 

Fig. \ref{park} illustrates the effectiveness of our backend, showing a significant improvement before and after the semi-pose-graph BA on the long walking sequence \emph{Park}. Over 2\,km of walking and revisits through parks and corridors, the estimated trajectory remains smooth and highly consistent, as evidenced by the clean building edges and park walls in the point cloud map.

\subsection{Performance on Hilti-Oxford Dataset}
\begin{table}[h]
\vspace*{-2mm}
\caption{ATE in meter and obtained scores on Hilti-Oxford Dataset.}
\vspace*{-4mm}
\label{hilti}
\begin{center}
\resizebox{0.48\textwidth}{!}{
\begin{tabular}{c|ccccccccc}
\toprule
\ & \ exp01 & \ exp02 & \ exp03 & \ exp07 & \ exp09 & \ exp11 & \ exp15 & \ exp21 & \ score\\
\midrule
OKVIS2 \ & 0.12 & 0.18 & 0.65 & \textbf{0.13} & \textbf{0.25} & \textbf{0.09} & 0.19 & 0.41 & 32.5 \\
Our VIO \ & 0.16 & 0.28 & 0.30 & 0.15 & 0.38 & 0.17 & 0.18 & 0.30 & 22.2 \\
Our SLAM \ & \textbf{0.10} & \textbf{0.15} & \textbf{0.15} & 0.14 & 0.28 & 0.21 & \textbf{0.17} & \textbf{0.30} & 40.9 \\
\bottomrule
\end{tabular}}
\vspace*{-4mm}
\end{center}
\end{table}
The Hilti-Oxford dataset \cite{zhang2022hilti} is part of the Hilti SLAM challenge 2022, and it features a sensor configuration similar to that of the Newer College dataset, consisting of five fisheye cameras and an IMU. To reduce the computational resources needed, we chose four views from five cameras for processing, omitting the top view as it has fewer features. The dataset provides millimeter-accurate groundtruth at sparse locations and a web interface for evaluation as well as a score leaderboard are provided for benchmarking. The RMSE ATE on sparse groundtruth is used as the evaluation metric, and the results are summarized in Tab. \ref{hilti}. Our VIO system achieved ATEs ranging from 10\,cm to 30\,cm. Exp01-exp03 were collected in a construction site with start and end points at the same place, allowing for loop closures. Thus our full-SLAM system shows improvements for these sequences, resulting in a higher score that outperforms the winner approach of Hilti SLAM challenge 2022 by OKVIS2.

\subsection{Processing Time}\label{sec:runtime}
\vspace*{-2mm}
\begin{table}[h!]
\vspace*{-3mm}
\caption{Processing time of different modules (Unit: Millisecond).}
\vspace*{-7mm} 
\label{runtime}
\begin{center}
\resizebox{0.49\textwidth}{!}{
\begin{tabular}{c|cccccccc}
\toprule
& \ \ \multirow{2}{*}{IO} \ & \ Extract \ & \ Check & Preintegrate & \ \multirow{2}{*}{Tracking} & \ \multirow{2}{*}{Local-BA} & \ Compute & \ \multirow{2}{*}{Total} \\
&& \ feature map \ & \ keyframe & IMU &&& \ rel. pose & \\
\midrule
Non-keyframe    & 7 & 4 & 2 & - & - & - & - & 13 \\
Keyframe        & 7 & 4 & 2 & 3 & 48 & 106 & 8 & 178 \\
\bottomrule
\end{tabular}}
\vspace*{-4mm}
\end{center}
\end{table}
The runtime performance of the system is evaluated on the TUM-VI dataset for both non-keyframe and keyframe cases. The results are presented in Tab. \ref{runtime}, which breaks down the time into different stages. For non-keyframes, the incoming image is processed up to the keyframe-check step. The average time taken for non-keyframes is 13 milliseconds. In contrast, each keyframe is further tracked by running four iterations single-frame BA followed by two iterations of local BA. Finally, the relative pose factors are computed for the keyframes and their associated reprojection factors outside current sliding window. On average, it takes around 175 milliseconds to process one keyframe.

The system achieves a good balance between processing speed and accuracy, selecting on average 23.5\% of the frames as keyframes while still processing an average of 25 frames per second on the TUM-VI dataset, exceeding the recording time by 25\%. However, in multi-camera mode, the increased computation reduces the processing speed to an average of 22 frames per second on the Newer College and Hilti-Oxford dataset. On the backend, loop closuring and global semi-pose-graph BA are carried out once by the end of program, which can take several seconds to a few minutes for long sequences such as the Newer College \emph{Park} sequence.

\section{CONCLUSION}
This paper presents a state-of-the-art visual-inertial SLAM system capable of fusing visual reprojection factors from multiple fisheye cameras and IMU preintegration factors into a unified factor graph. To achieve globally consistent mapping, a novel semi-pose-graph bundle adjustment scheme has been proposed to replace the computationally expensive full BA. By running alternative BA optimizations and RFT predictions, the system can converge to the maximum likelihood solution. Extensive evaluations on multiple datasets validate the accuracy and robustness of our proposed system. As part of future work, we aim to enhance the robustness in large-scale outdoor scenes, where noisy pixels like sky and shadows should be better handled. Furthermore, we plan to incorporate a global loop closure detection method, such as Bag-of-Word, to strength the loop closing ability in case of large odometry drift caused by long-distance travel.

\addtolength{\textheight}{-3cm}   






\bibliographystyle{IEEEtran}
\bibliography{IEEEabrv,mybibfile}

\end{document}